\documentclass[amsmath,amssymb,aps,superscriptaddress,reprint]{revtex4-1}
\usepackage{times}
\usepackage{graphicx}%
\usepackage{mathrsfs}%
\usepackage{xcolor}%

\begin{document}

\title{Flashpoints Signal Hidden Inherent Instabilities in Land-Use Planning}

\author{Hazhir \surname{Aliahmadi}}
\affiliation{Department of Physics, Engineering Physics, and Astronomy, Queen's University, Kingston ON, K7L 3N6, Canada}

\author{Maeve \surname{Beckett}}
\affiliation{Department of Physics, Engineering Physics, and Astronomy, Queen's University, Kingston ON, K7L 3N6, Canada}

\author{Sam \surname{Connolly}}
\affiliation{Department of Physics, Engineering Physics, and Astronomy, Queen's University, Kingston ON, K7L 3N6, Canada}

\author{Dongmei \surname{Chen}}
\affiliation{Department of Geography and Planning, Queen's University, Kingston ON, K7L 3N6, Canada}

\author{Greg \surname{van Anders}}
\affiliation{Department of Physics, Engineering Physics, and Astronomy, Queen's University, Kingston ON, K7L 3N6, Canada}

\begin{abstract}
  Land-use decision-making processes have a long history of producing globally
  pervasive systemic equity and sustainability concerns. Quantitative,
  optimization-based planning approaches, e.g.\ Multi-Objective Land Allocation
  (MOLA), seemingly open the possibility to improve objectivity and transparency
  by explicitly evaluating planning priorities by the type, amount, and location
  of land uses. Here, we show that optimization-based planning approaches with
  generic planning criteria generate a series of unstable ``flashpoints''
  whereby tiny changes in planning priorities produce large-scale changes in the
  amount of land use by type. We give quantitative arguments that the flashpoints
  we uncover in MOLA models are examples of a more general family of
  instabilities that occur whenever planning accounts for factors that
  coordinate use on- and between-sites, regardless of whether these planning
  factors are formulated explicitly or implicitly.  We show that instabilities
  lead to regions of ambiguity in land-use type that we term ``gray areas''. By
  directly mapping gray areas between flashpoints, we show that quantitative
  methods retain utility by reducing combinatorially large spaces of possible
  land-use patterns to a small, characteristic set that can engage stakeholders
  to arrive at more efficient and just outcomes.
\end{abstract}

\maketitle

\section{Introduction}\label{sec1}

Sustainable, equitable land use will be a major challenge in this century due to
the effects of urbanization \cite{seto_global_2012,unWUP2018}, climate change
\cite{PrecipitationExtremes,winsemiusGlobalDriversFuture2016,DrylandExpansion}, and renewable energy
generation \cite{van_de_ven_potential_2021}. Historically, allocating land-use
qualitatively has produced outcomes that have raised serious, continuing issues
regarding sustainability \cite{bibri_compact_2020} and
equity\cite{anguelovski_equity_2016}. Inequitable and unsustainable land use can
arise through planning processes that are top-down, subjective, and opaque.
These shortcomings suggest a step forward through decision-making processes that
explicitly and transparently articulate planning criteria for
land use by type, amount, and location. These land-use factors can be modelled
quantitatively, and optimized via techniques such as Multi-Objective Land
Allocation (MOLA) \cite{MOLA-Review}. However, although MOLA models for
land-use planning problems make criteria clear and precise, solving MOLA and
similar models typically relies on the use of optimization techniques
\cite{PS-MOLA,NGSA-MOLA, GA-MOLA-2004,MA-PS-MOLA,
NGSAII-MOLA-2015,MACO-MOLA,MOLA-SDG} in which the relationship between the
priorities of planning criteria and the resulting land-use allocations is
obscured. Obscuring priority--outcome relationships undermines the promise of
process-transparency that explicit models potentially provide for stakeholders.
Moreover, this obscurity of optimization methods could mask dangerous instabilities
in the models themselves. If that were the case,
a standard, optimization-supported land-use planning approach would not just be
unclear, it would also be unreliable.

Here, we show that common land-use planning criteria produce outcomes that are
extremely sensitive to the relative priority (weighting) of the criteria. We
analyze a MOLA model \cite{SongChenMOLA} based on the two most commonly used
planning criteria: compactness and suitability \cite{MOLA-Review},
and show that it exhibits a series of discrete, large-scale changes in land-use
allocation outcomes in response to minute changes in planning priorities. This
creates a series of instabilities we term ``flashpoints''. 

For each flashpoint, we compare the land-use patterns on either side and map
the discrepancies between the patterns. These discrepancies create a set of
locations we term ``gray areas'' that are regions of land-use that
are sensitive to infinitesimal changes in planning priorities. 

By comparing with mathematically identical models of magnetic materials, we find
that land-use flashpoints are equivalent to the most abrupt state changes known
in nature. This mathematical equivalence, coupled with the general understanding
developed over several decades of the study of phenomena in magnetism, indicates
that an extremely broad class of MOLA models will exhibit flashpoints of the
type we observe. This generality is driven by a conflict between the
coordination of use between multiple land parcels and the coordination of a
parcel's use with preexisting human or natural factors. These results
indicate that conventional approaches to optimization-supported land-use
planning are unreliable.

Though the inherent instability of MOLA means that there is no globally optimal
land-use pattern for all suitability priority levels, our analysis yields
a distribution of patterns that provides insight into the underlying trade-offs
between planning priorities. We use techniques from statistical physics to
construct explicit relationships between sets of planning priorities and the
resulting land-use allocations they produce. By formulating priority--outcome
relationships explicitly, we reduce the large space of possible land-use
outcomes to a small set of representative patterns that are reliable over
identifiable ranges of planning priorities. Explicit priority--pattern
relationships of the form we construct here restore transparency to explicit
models, and provide concrete information about planning trade-offs that can
empower stakeholders to engage in conversation that could lead to more
sustainable and equitable land-use decision making.

\section{Results}\label{sec2}
\begin{figure*}
  \includegraphics[width=1.0\textwidth]{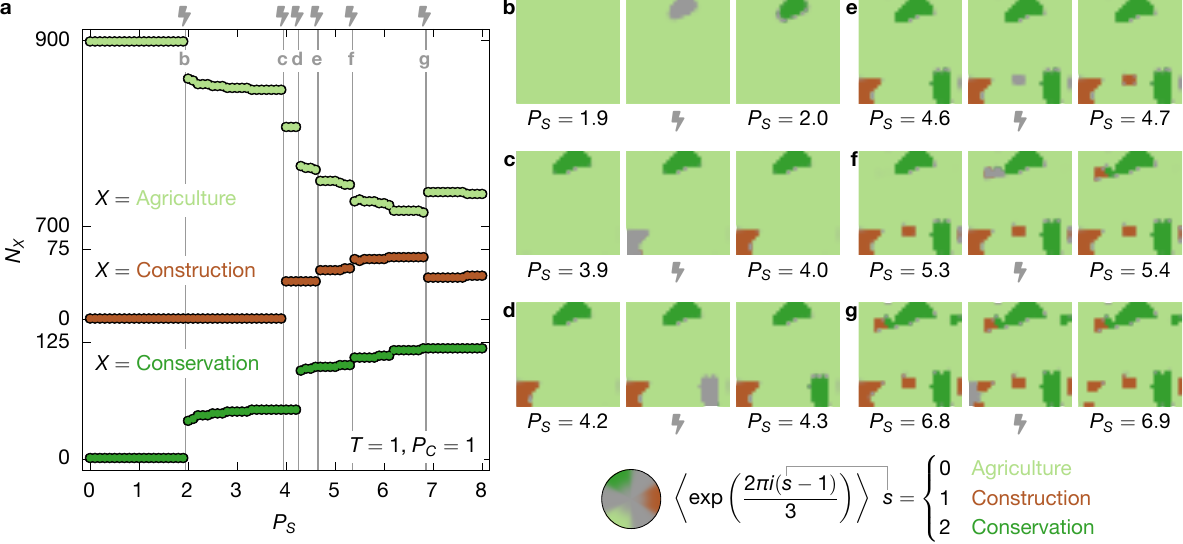}
  \caption{
    Changing priorities in quantitative land-use allocation generates flashpoints: priority
    sets in which small changes generate large-scale reorganization in land use
    patterns. \textbf{a} The optimal land-use fraction computed via simulated annealing-like 
    methods with low annealing threshold ($T=1$) as a function of
    suitability priority level $P_S$. With changing priority, the land-use
    fraction shows relatively stable regions punctuated by a series of discrete
    jumps. \textbf{b-g} Comparing land-use patterns on either side of each
    flashpoint shows that areas of change are generally clusters of parcels
    rather than peripherals or borders between land-use types. Maps are computed
    by labelling each land-use type with an integer, transforming it to a point on
    a circle in the complex plane, and then averaging over hundreds or thousands
    of statistically independent realizations of land-use patterns, according to
    the key below panels \textbf{b-g}.
  }
  \label{Fig-Flashpoints}
\end{figure*}

\textbf{Compactness-suitability trade-off produces flashpoints.} 
To determine the dependence of globally optimal land-use distributions on the
choice of priorities, we analyzed land-use patterns from 82,500 statistically
independent simulations of the multi-objective land allocation model from Ref.\
\cite{SongChenMOLA} for various levels of suitability priority relative to
compactness. Results are shown in Fig.~\ref{Fig-Flashpoints}.

In principle, land-use patterns could depend continuously or discontinuously on
the priority of the suitability criterion, $P_S$. We determine this dependence
in Fig.~\ref{Fig-Flashpoints}a, which shows the globally optimal land-use pattern
in terms of the land-use fraction of land-use types. We aggregated an average
of $2\times 10^5$ land-use patterns for each suitability priority. We
determined global optima from inferred optimization landscapes, which we
computed via Landau free energy minima (see Methods), using standard
accumulation techniques, e.g.\ \cite{entint}. Apart from minor continuous
changes in the land-use distribution for $2.0 < P_S < 3.9$ and $5.4 < P_S <
6.8$, the response to changes of suitability priority was characterized by a
series of abrupt, discontinuous changes we refer to as flashpoints (see Methods
for cutoff).

\noindent\textbf{Flashpoint instabilities produce gray areas.} 
According to the criteria outlined in Methods, we identified six flashpoints
between $0.1 < P_S < 8$. The existence of a flashpoint signals a significant
redistribution of the land-use pattern in response to a small change in
suitability priority. The corresponding redistribution at a flashpoint will
produce regions of uncertainty in land use. Fig.\ \ref{Fig-Flashpoints}b-g, 
middle images, depict these regions, called ``gray areas''.

We determined gray areas by aggregating land-use patterns at Landau free energy
minima. We represent each parcel's land-use type by a point on a circle where
the three land-use types are separated by 120 degrees (see Methods). Taking the
average of those points across several thousands of land-use patterns gives an
average for each parcel within the circle, as shown in the bottom right of Fig.\
\ref{Fig-Flashpoints}. Suitability priority just below a flashpoint (Fig.\
\ref{Fig-Flashpoints}b-g, left images) or above that (right images) results in a
consistent land-use pattern, i.e.,\ little to no gray, across samples. However,
comparing land-use patterns on either side of the flashpoints shows significant
gray areas associated with different land-use patterns. This finding indicates
that there are large regions with extreme parametric sensitivity.
\begin{figure*}
  \includegraphics[width=1.0\textwidth]{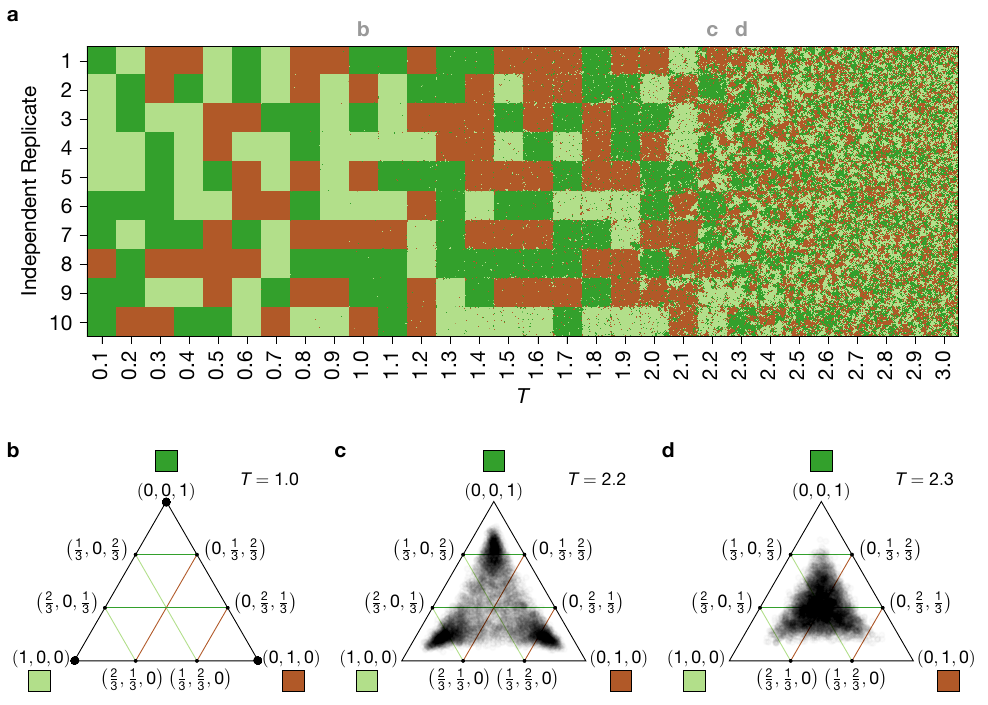}
  \caption{
    Compactness-only land-use allocation model shows spontaneous symmetry
    breaking with decreasing annealing threshold, indicating the underlying
    optimization landscape is rough.
    \textbf{a} Selected annealing replicates at fixed, decreasing annealing
    threshold (corresponding to $T$; decreasing from right to left) show
    compactness-only models at low annealing thresholds are characterized by a
    predominance of uniform, single land-use patterns. Data at each threshold
    are shown for 10 statistically independent annealing replicates. Data from
    all replicates are aggregated in ternary plots (see Methods) in SI Movie 2
    and panels \textbf{b}-\textbf{d}. Panels \textbf{b}-\textbf{d} show land-use
    where the vertices of the triangle correspond to single-use patterns. Each
    simulation sample is plotted as an ordered triple, representing land-use
    fractions of land-use types (Agriculture, Construction, Conservation).
    Coloured lines indicate lines of constant land-use fractions of the
    corresponding land-use type, and the centroid of the triangle represents an
    equal distribution of land-uses among the three types. Panel \textbf{b}
    shows that at a low threshold ($T=1$) data points localize the ternary
    vertices, indicating near single-use maps. In contrast, at a higher threshold
    ($T=2.3$; panel \textbf{d}), data localize with triangular symmetry at the
    centroid, signalling unbroken symmetry. However, at a slightly
    lower threshold ($T=2.2$; panel \textbf{c}) land-use patterns are not
    concentrated about the centroid but are instead becoming peripheral,
    signalling that so-called spontaneous symmetry breaking underlies the rough
    optimization landscape.
  }
  \label{Fig-BrokenSymmetry}
\end{figure*}

\noindent\textbf{Spontaneous Symmetry Breaking Creates the Conditions for
Flashpoints.} To understand how flashpoints arise, we consider a simplified
model with a vanishing suitability priority, $P_S=0$, so that land-use patterns
optimize spatial compactness alone. In the analysis above, we worked at fixed
annealing threshold, $T$. Now, instead we vary the threshold and track emergent
land-use patterns as a function of the threshold. Results for the
compactness-only model are depicted in Fig.~\ref{Fig-BrokenSymmetry}.
Fig.~\ref{Fig-BrokenSymmetry}a shows randomly selected land-use snapshots at a
range of annealing thresholds.  This compactness-only model shows non-compact
solutions at high thresholds, while land-use patterns become compact at low
thresholds, with one land-use type dominating the entire map. Prevailing
single-use patterns at low annealing thresholds, in a model that makes no
explicit preferential distinction between land-use types, occur because multiple
local minima emerge corresponding to each possible land use. The emergence of
multiple local minima signals a phenomenon termed spontaneous symmetry breaking
\cite{goldenfeld} in physics.

Fig.~\ref{Fig-BrokenSymmetry}b plots distributions of all sampled land-use
patterns at thresholds corresponding to $T=1.0$, indicating broken
symmetry with minimal fluctuations, validating our choice of $T=1.0$
in the flashpoint analysis shown in Fig.~\ref{Fig-Flashpoints}. A similar
analysis of land-use distributions is shown in Fig.~\ref{Fig-BrokenSymmetry}c-d
suggests symmetry breaking occurs between $2.2\le T\le 2.3$. Results for all
thresholds are shown in Supplementary Movie 2.

\begin{figure*}
  \includegraphics[width=\textwidth]{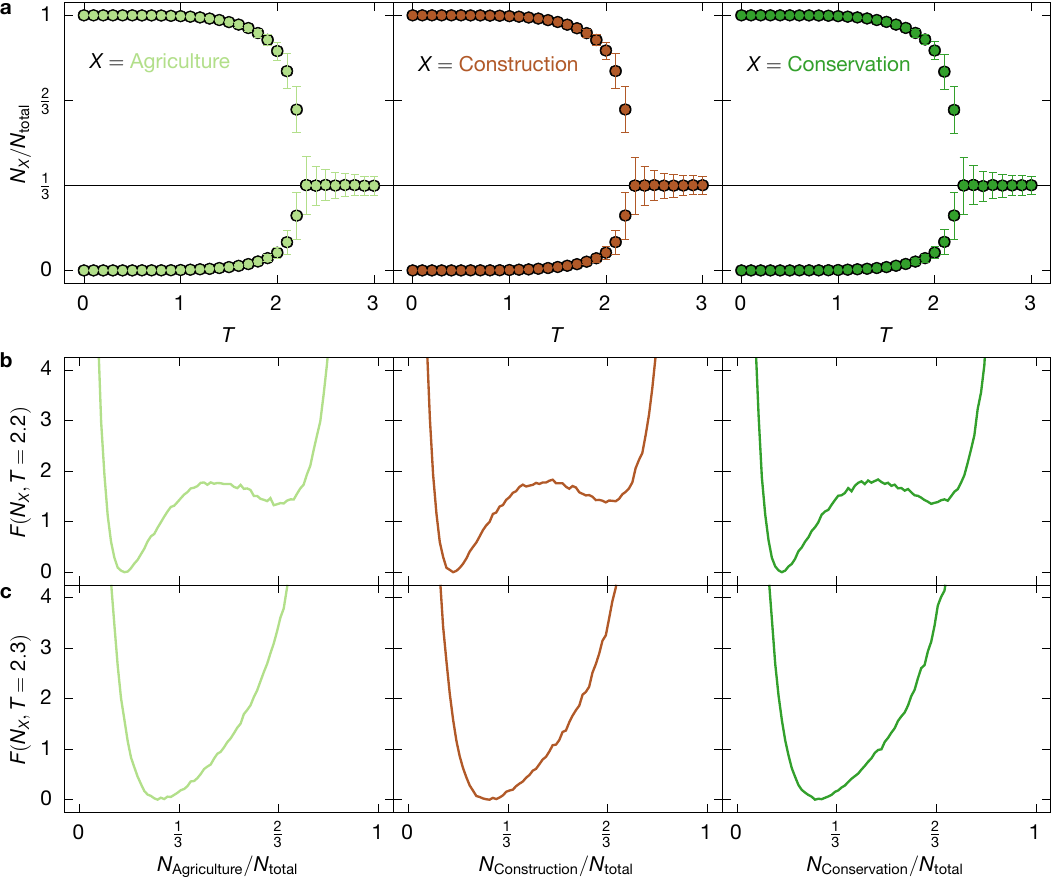}
  \caption{
    Inferred optimization landscapes (Landau free energy) at vanishing
    suitability priority confirms spatial compactness drives spontaneous
    symmetry breaking. \textbf{a} Landau free
    energy minima at a range of thresholds indicates low thresholds
    spontaneously fix single land use patterns. Data points indicate positions
    of free energy minima extracted from sampling more than 2,000 statistically
    independent simulations spread across the indicated range of annealing
    thresholds. \textbf{b,c} Landau free energy computed as a function of land
    use fraction confirms symmetry breaking occurs at a threshold $2.2\le T\le
    2.3$. Statistical error is on the order of the ``jitter'' in the data
    traces.
  }
  \label{Fig-LandauFE}
\end{figure*}

To confirm that spontaneous symmetry braking is occurring, we computed inferred
optimization landscapes, via Landau free energies (see Methods), as a function
of land-use fraction for each studied threshold. Fig.~\ref{Fig-LandauFE}a shows
locations of free energy minima as a function of threshold, and panels b and c
show results at $T=2.2$ and $2.3$ confirming spontaneous symmetry breaking
occurs between those thresholds.  The increase in the statistical uncertainty
of the land-use fraction near the symmetry-breaking threshold is a known
physical effect that comes from diverging fluctuation magnitudes near a
so-called critical point \cite{goldenfeld}.

Comparing the compactness-only model with the compactness--suitability trade-off
model suggests that increasing suitability priority at low annealing-thresholds
disrupts ``local'' balances between compactness and suitability. When local
balance is disrupted, the result is a discrete, large-scale use reallocation of
many land parcels in a region between single-use and multi-use patterns.

\noindent\textbf{Flashpoints from Generic Disrupted Multi-site/On-site
Coordination Balance.} 
It is possible to substantiate this disrupted compactness--suitability balance
argument quantitatively, and more generally. Generic planning objectives can
arise from human factors (e.g.\ social, economic, or demographic) and
natural factors (e.g.\ geographic, ecological, environmental).  Both human and
natural factors can drive the land use of a parcel to either coordinate with
multiple adjacent parcels or to coordinate with the features or properties of
the parcel itself \cite{MOLA-Review}. Examples of drivers of multi-site
coordination include the provision of social or community services and
infrastructure, \cite{duhKnowledgeinformedParetoSimulated2007a}; or contiguity of
habitat for wildlife. Examples of drivers of on-site coordination include prior
land use, flood risk, terrain, soil quality, etc.

To formulate the balance between multi-site and on-site coordination
quantitatively, note that, generically, multi-site coordination will drive
spontaneous symmetry breaking that will induce regions with homogeneous,
coordinated use. For parcels in a region $R$ of the map, the objective 
``cost'' of reallocation of the land use is given by
\begin{equation}
  \Delta F = \Delta E_{M} L(\partial R) - \Delta E_{O} A(R) \; ,
  \label{eq:DF}
\end{equation}
where $\Delta E_{M}$ is the average objective increase for breaking multi-site
coordination per unit length of the boundary, $\partial R$, of region $R$, $L$
is the length of the boundary, $\Delta E_{O}$ is the average objective reduction
for on-site coordination per unit area, and $A$ is the area of the region. A
region will switch its  land-use if $\Delta F<0$, i.e.\ if it reduces overall objective
cost.  Reducing objective cost by breaking homogeneity in a region will occur
whenever the average on-site cost reduction is sufficient that
\begin{equation}
  \Delta E_{O} > \frac{\Delta E_{M} L(\partial R)}{A(R)} \; .
  \label{eq:FlipThresh}
\end{equation}
In this relationship, which is an application of general principles of
nucleation theory \cite{debenedetti_metastable_1996}, the left-hand side scales
linearly with the priority for on-site coordination, whereas the right-hand side
scales linearly with the priority for multi-site coordination. For a region of
any geometry defined by a given $L/A$, there will be some set of weights
that cross the condition defined in Eq.\ \eqref{eq:FlipThresh}. Thus flashpoints
are inherent in any land allocation problem with allocation objectives that
involve multi-site and on-site coordination, regardless of the details of the
optimization method or the model.

\section{Discussion}
We analyzed several tens of thousands of statistically independent
multi-objective land allocation patterns spread over a range of planning
priorities. We found optimal land-use amounts by type exhibited extreme
sensitivity to the priorities of planning criteria, with use type undergoing a
series of discrete, large-scale changes we termed flashpoints. Flashpoint
instabilities signal that optimization-based modelling like MOLA, conventionally
deployed, is not a reliable support for land-use planning. We mapped
corresponding spatial locations of land-use instability, called gray
areas.

Our results were produced using simulated annealing-like methods applied to a
specific model that is known in the literature, however, the effects we found are
more general. We compared allocation models with mathematically analogous
systems that model the physics of magnetism. The mathematical equivalence
between quantitative land-use models and a broad class of magnetic systems for
which a consistent understanding developed through decades of theoretical,
computational, and experimental investigation with multiple approaches
\cite{chikazumi_physics_2009} means that the instabilities we observed are not
artifacts of our specific model or methods. Instead, our results signal that
flashpoints are a generic feature of any land allocation problem that combines
objectives for compactness and suitability.

\begin{figure*}
  \includegraphics[width=\textwidth]{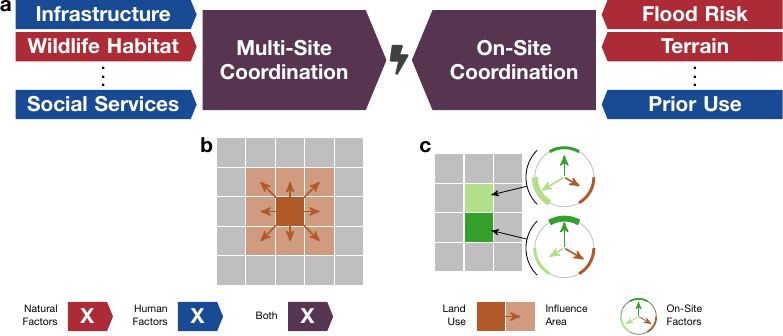}
  \caption{
    Land-use flashpoints arise generically when
    common human and natural
    considerations drive conflict between multi-site land use coordination and
    on-site coordination. \textbf{a} Both human factors and
    natural factors induce planning considerations about the
    coordination between adjacent land parcels and coordination of a land
    parcel's use with the, possibly pre-existing, properties. Multi-site and
    on-site forms of coordination combine to generate the potential for
    instability that produces flashpoints. \textbf{b} Multi-site coordination
    arises because a given land use for a particular parcel generates an
    influence area on adjacent parcels. \textbf{c} On-site coordination
    drives the use of a parcel according to local factors.
  }
  \label{Fig-collisionmodel}
\end{figure*}

To relate our findings to other methods for similar problems, it is important to
note that the approach we took to MOLA here maps the parameters of the model to
a statistical distribution of candidate solutions. This distribution is
determined by the structure of the underlying solution space, so it is a
property of the land-use model, not the optimization method. Other methods such
as genetic algorithms
\cite{GA-MOLA-2004,NGSAII-MOLA-2015,NGSA-MOLA,MOLA-SDG}, particle swarm
optimization \cite{PS-MOLA,MA-PS-MOLA}, or ant colony
optimization \cite{MACO-MOLA} might explore the solution space differently, but
that space would still be marked by the flashpoint instabilities we reported
here.
To relate our findings to other models, note that although the model we
used here had particular forms for multi-site and on-site coordination,
mathematically similar forms of coordination result from disparate human 
or natural factors.
For example, factors driving multi-site coordination such as
the importance of contiguity in wildlife habitats or the benefits of contiguous
communities for providing infrastructure or public services suggest spatial
compactness in a generic form similar to the present model.
Indeed, a recent
systematic review found that compactness is the most common MOLA-modelling
criterion \cite{MOLA-Review}.  The same systematic review has also shown that
suitability is a frequent model criterion \cite{MOLA-Review}, but any other
factor that produces on-site coordination, e.g., distributions of prior use,
will lead to similar effects. Some form of spatial inhomogeneity is also
inevitable since regions of land frequently have differences in, e.g., flood
risk, soil quality, proximity to water, or terrain. The interaction of
multi-site and on-site coordination will generally produce flashpoints (see
Fig.\ \ref{Fig-collisionmodel}). 

Our investigation indicates that, generically, land-use allocation models reduce
the large space of possible land-use patterns to a small set of basic
arrangements that are punctuated by extreme sensitivity to the relative priority
of competing planning criteria. Simply making a choice about the priority of
various objectives and then determining an optimal solution will produce results
that can exhibit spectacular dependence on arbitrarily small changes in planning
priorities. In quantitative models, the abruptness of these changes in
optimization outcomes is mathematically equivalent to so-called first-order
phase transitions, the most abrupt changes of state that are known in
nature \cite{Sethna2021}. Abruptly triggered instabilities run directly
counter to the resilience that has long been sought in managing the link between
social and ecological systems \cite{berkes_linking_2000,defries_land-use_2004}.
\begin{figure*}
  \includegraphics[width=1.0\textwidth]{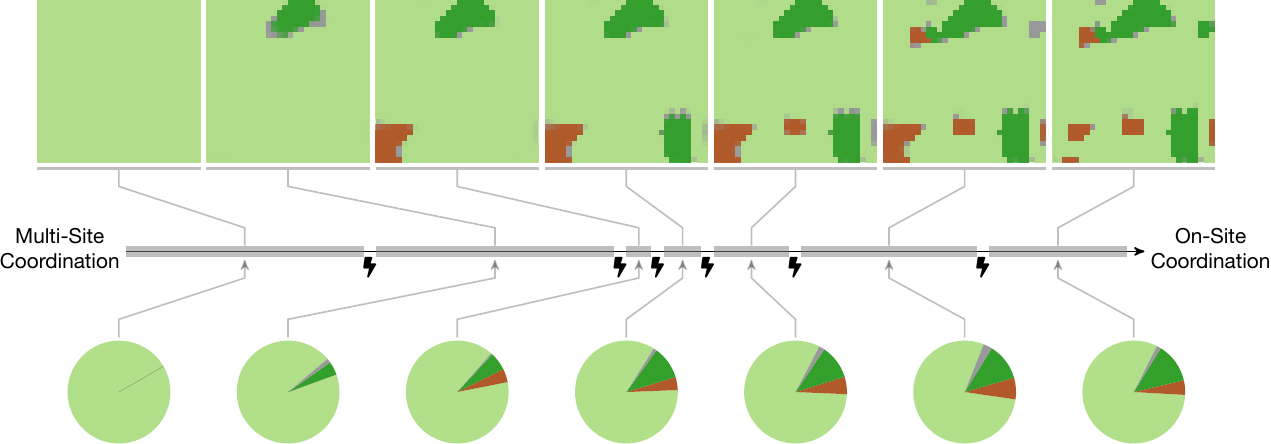}
  \caption{
    Substantive science--policy dialogue requires charting explicit connections
    between planning goals (pie charts, bottom), planning priorities (axis,
    middle), and planning outcomes (maps, top). We match pie charts showing
    possible targets for fractions of land allocated for given uses to
    corresponding families of land-allocation models that differ in planning
    priorities but yield similar allocations. We further match fractional
    allocations with spatial distributions of land-use types. Colour shading
    indicates regions of commonality across a family of models, and gray shading
    indicates variation.
  }
  \label{phase_diagram}
\end{figure*}

Our analysis of land-use change was quantitative. However, generic features of
the broad class of models that are similar to ours indicate that land-use
decision-making that confronts multi-site and on-site coordination is subject to
a perilous balance regardless of whether these factors are explicitly quantified 
or not. Rather than using quantitative approaches to land-use allocation for
top-down optimization, our analysis suggests quantitative approaches could more
profitably confront an unstable balance among planning priorities by charting
relationships between the priorities and land-use outcomes (see Fig.\
\ref{phase_diagram}).

Finally, top-down, allocation modelling alone cannot drive sustainable land-use
planning. In land-use planning, as in other domains, sustainability requires
iterative, reciprocal interaction that drives coproduction between science and
policy \cite{clark_crafting_2016}. Our approach expands allocation techniques
because it moves beyond singular, putative optimal outcomes of a given model,
and renders explicit priority--outcome relationships among families of
underlying land-use models. Making these modelling relationships explicit is a
critical step forward in nurturing conversations between scientists and policymakers that inform refined or revised models to better support sustainable
land-use planning.

\section{Methods}
\subsection{Multi-Objective Land Allocation Model}
We model land-use patterns as a set of discrete parcels each with a potentially
different use subject to overall planning criteria following established
conventions for multi-objective land allocation (MOLA) models
\cite{MOLA-Review}. Systematic review \cite{MOLA-Review} has shown that spatial
compactness and suitability are generic MOLA planning criteria. For
concreteness, we work with a model described in Ref.\ \cite{SongChenMOLA} with
compactness and suitability criteria that are sufficiently representative to
facilitate extracting generalizable lessons. The data used for this study are based on a semi-hypothetical dataset provided in \cite{NGSA-MOLA}, which pertains to a square area of 9 km$^2$ located in the Xin’andu township of Dongxihu District, Wuhan, China.

Following Ref.\ \cite{SongChenMOLA}, we take the criterion for compactness as
\begin{equation}
\label{eqn:Compactness}
O_{1}=-\sum_{s=0}^{S}\sum_{i=1}^{N}\sum_{j=1}^{M} b_{ijs} x_{ijs} \; ,
\end{equation}
where $M$ and $N$ are the number of columns and rows of the map, $S$ is the
number of land-use types, and
$x_{ijs}$ is $1$ if the parcel at $(i,j)$ is allocated for land-use $s$, and $0$
otherwise. The coefficient $b_{ijs}$ counts the number matching neighbours for
the $(i,j)$ parcel
\begin{equation}
b_{ijs} = x_{i+1 j+1 s} + x_{i j+1 s} + x_{i+1 j s} + x_{i-1 j s}
+ x_{i j-1 s} + x_{i-1 j-1 s} + x_{i-1 j+1 s} + x_{i+1 j-1 s} \; ,
\end{equation}
We take suitability criterion as
\begin{equation}
\label{eqn:Suitability}
O_{2}=-\sum_{s=0}^{S}\sum_{i=1}^{N}\sum_{j=1}^{M} c_{ijs} x_{ijs} \; ,
\end{equation}
where $c_{ijs}$ parametrizes the suitability of the $(i,j)$ parcel for land-use
type $s$. Here, we follow Ref.\ \cite{SongChenMOLA} using $M=N=30$ to give a $30
\times 30$ map, and three land-use types; agriculture, construction, and
conservation labelled with the integers $s=0,1,2$, respectively. Suitability
data $c_{ijs}$ are given in SI.

We form a weighted combination of the planning criteria Eqs.\
\eqref{eqn:Compactness} and \eqref{eqn:Suitability} as a Hamiltonian
\begin{equation}
\label{eqn:Hamiltonian}
H = P_C O_{1} + P_S O_2 \; ,
\end{equation}
where $P_C$ and $P_S$ are weights that correspond to priorities for the compactness
and suitability objectives, and are physical analogues of pressure \cite{systemphys}.
$P_C \gg P_S$ or $P_C \ll P_S$ generate allocations dominated by one criterion
or another. Trade-offs occur when $P_C$ and $P_S$ are of a similar order. We fixed
units by setting $P_C=1$ and taking $0.1<P_S<8$, which covers the essential
region of compactness/suitability trade-offs. Similar to techniques from
simulated annealing \cite{kirkpatrick1983}, we sample patterns using a Markov
Chain Monte Carlo algorithm.

\subsection{Markov Chain Monte Carlo Sampling}
We study the multi-objective land allocation model using Markov Chain Monte
Carlo (MCMC) techniques, using the Wolff algorithm.\cite{wolff} The presence of
suitability objective creates a spatial inhomogeneity that violates a symmetry
assumption of the standard Wolff algorithm, so we implemented a modified version
described in Ref.\ \cite{GhostWolff}. Land-use patterns are sampled according to
an annealing threshold $k_B T$, where $T$ is equivalent to the temperature in
simulated annealing approaches to MOLA (e.g.\ Ref.\ \cite{SongChenMOLA}), and
$k_B$ is a constant expressing the conversion between energy and temperature
units (we set $k_B=1$ without loss of generality). C++/Python source code is
available at Ref.\ \cite{connollyFlashWolff2023}.

To ensure accurate results, for our flashpoint analysis we thermalized systems
of $30\times 30$ land units at a temperature of $T=15$ for $10^3$ Monte
Carlo (MC) sweeps. We cooled systems to a target temperature of $T=1$ over
$3.5\times 10^4$ MC sweeps. We equilibrated systems at the target temperature
for $10^4$ MC sweeps. We measured land-use distributions at intervals of $50$ MC
sweeps over $10^4$ sweeps. We computed an average of more than $10^3$
statistically-independent simulations over the entire suitability priority range
using a minimum of $300$ random seeds at priorities far from flashpoints, and
many more near flashpoints. Data are available at Ref.\
\cite{aliahmadi_land_2023}.

\subsection{Ternary Plots}
Our MC sampling approach produced more than $1.6\times 10^7$ land-use patterns.
A complete display in the form of Fig.\ \ref{Fig-BrokenSymmetry}a would require
more than $1.4\times 10^{10}$ pixels. We, therefore, analyzed land-use patterns
by aggregating land use of parcels across each pattern for each land-use type
($N_X$ where $X$ represents either agriculture, construction, or conservation).
We represented each pattern as an ordered triple. Considering ordered triples
as defining a three-dimensional space, ordered triples for fixed map size all
lie on a plane that is normal to the vector $(1,1,1)$. We projected our ordered
triples to this plane, where all possible land-use patterns fall inside an
equilateral triangle, with single-use patterns at the vertices.

We plotted land-use totals by placing a marker for observed patterns in
Fig.\ \ref{Fig-BrokenSymmetry}b-d, and in SI Movies 1 and 2. In order to not
violate the limits of our plotting software, we plotted $5000$ randomly selected
patterns at each choice of the priority set for a fixed annealing threshold.

\subsection{Landau Free Energy}
In generic multi-objective, non-convex optimization problems, possible outcomes
depend non-trivially on system configurations and on the relative priority of
objectives. To determine the form of the outcome landscape in physical systems
it is conventional to infer its form from statistical sampling where the
landscape is known as the Landau free energy \cite{goldenfeld}.

We determine optimal land use patterns for a given set of planning priorities
by generating patterns via MCMC sampling as described above. We aggregated
sampled patterns by the count of uses of each type. We accumulated these counts
over thousands of samples to compute a histogram by land-use type. This
histogram produces a Monte Carlo estimate of the probability of sampling
patterns of fixed-use counts. The negative logarithm of this distribution gives
the inferred optimization landscape (Landau free energy \cite{goldenfeld}), and
the global minimum of this landscape corresponds to the optimal land use pattern.

Additionally, the structure of a Landau free energy surface as a function of
system parameters encodes distinct regimes of system behaviour and the
abruptness of transitions between them \cite{goldenfeld}. Continuous, so-called
symmetry-breaking, transitions occur when changes in system parameters induce a
single, global minimum to split into multiple minima \cite{goldenfeld}, e.g.,
as seen in Fig.\ \ref{Fig-LandauFE}. Abrupt, so-called first-order, transitions
occur when changes in system parameters induce an exchange between local and
global minima \cite{goldenfeld}, e.g., which drives the behaviour in Fig.\
\ref{Fig-Flashpoints}.

\subsection{Flashpoint Cutoff}
A flashpoint is an abrupt change in land use outcomes in response to slight
changes in planning priorities. For this investigation, we define a flashpoint
as a 10\% change in any land-use type across a single incremental change in
suitability priority according to
\begin{equation}
\label{FlashpointTh}
\frac{|N_X(P_S^{n+1})-N_X(P_S^{n})|}{\tfrac{1}{2}[N_X(P_S^{n+1})+N_X(P_S^{n})]} > \alpha
\end{equation}
where $N_X(P_S^{n})$ is the number of land use $X$ at the n\textsuperscript{th}
$P_S$, and $\alpha=0.1$ is the flashpoint cutoff.

To ensure accurate $N_X$ we computed Landau Free Energy minima \cite{goldenfeld}
at $T=1.0$, sufficiently low that thermal fluctuations are small. Lower
temperatures would only sharpen the behaviour we observed. We averaged results
from 500 statistically independent replicates for each $0.1<P_S<8.0$ in
increments of $0.1$. For priorities near the flashpoints, we used 2000
replications to ensure accuracy.

\subsection{Gray Areas}
We mapped gray areas via a quantity that expresses local
land-use patterns. We did that by labelling each parcel of a pattern by an
integer $s=0,1,2$, corresponding to the land-use types; agriculture,
construction, and conservation, respectively.  Following the Landau free energy
computation for the flashpoint threshold, described above, we aggregate patterns
at the free energy global minimum and map each parcel's land-use type to
a point on the unit circle in the complex plane via
\begin{equation}
  z = \exp\left(\frac{2\pi i (s-1)}{3}\right) \; .
  \label{eq-zmap}
\end{equation}
We then average $z$ of each parcel at the free energy minimum for the
suitability priority immediately to either side of the flashpoint, and the
combination of both sides. This average $\left<z\right>$ over the set of mapped
points gives a point on the unit circle for each parcel, which we shade
according to the colour map shown in Fig.\ \ref{Fig-Flashpoints}.

\section{Acknowledgements}
We acknowledge the support of the Natural Sciences and Engineering Research
Council of Canada (NSERC) grants RGPIN-2019-05655, DGECR-2019-00469, and
RGPIN-2019-05773.  Computations were performed on resources and with support
provided by the Centre for Advanced Computing (CAC) at Queen's University in
Kingston, Ontario. The CAC is funded by: the Canada Foundation for Innovation,
the Government of Ontario, and Queen's University. GvA acknowledges the
hospitality of the Kavli Institute for Theoretical Physics (KITP) where this
work was completed. This research was supported in part by the National Science
Foundation under Grant No. NSF PHY-1748958. We also thank Dr.\ Mingjie Song for
providing the land use and suitability data.


\begin{thebibliography}{30}%
\makeatletter
\providecommand \@ifxundefined [1]{%
 \@ifx{#1\undefined}
}%
\providecommand \@ifnum [1]{%
 \ifnum #1\expandafter \@firstoftwo
 \else \expandafter \@secondoftwo
 \fi
}%
\providecommand \@ifx [1]{%
 \ifx #1\expandafter \@firstoftwo
 \else \expandafter \@secondoftwo
 \fi
}%
\providecommand \natexlab [1]{#1}%
\providecommand \enquote  [1]{``#1''}%
\providecommand \bibnamefont  [1]{#1}%
\providecommand \bibfnamefont [1]{#1}%
\providecommand \citenamefont [1]{#1}%
\providecommand \href@noop [0]{\@secondoftwo}%
\providecommand \href [0]{\begingroup \@sanitize@url \@href}%
\providecommand \@href[1]{\@@startlink{#1}\@@href}%
\providecommand \@@href[1]{\endgroup#1\@@endlink}%
\providecommand \@sanitize@url [0]{\catcode `\\12\catcode `\$12\catcode
  `\&12\catcode `\#12\catcode `\^12\catcode `\_12\catcode `\%12\relax}%
\providecommand \@@startlink[1]{}%
\providecommand \@@endlink[0]{}%
\providecommand \url  [0]{\begingroup\@sanitize@url \@url }%
\providecommand \@url [1]{\endgroup\@href {#1}{\urlprefix }}%
\providecommand \urlprefix  [0]{URL }%
\providecommand \Eprint [0]{\href }%
\providecommand \doibase [0]{http://dx.doi.org/}%
\providecommand \selectlanguage [0]{\@gobble}%
\providecommand \bibinfo  [0]{\@secondoftwo}%
\providecommand \bibfield  [0]{\@secondoftwo}%
\providecommand \translation [1]{[#1]}%
\providecommand \BibitemOpen [0]{}%
\providecommand \bibitemStop [0]{}%
\providecommand \bibitemNoStop [0]{.\EOS\space}%
\providecommand \EOS [0]{\spacefactor3000\relax}%
\providecommand \BibitemShut  [1]{\csname bibitem#1\endcsname}%
\let\auto@bib@innerbib\@empty
%</preamble>
\bibitem [{\citenamefont {Seto}\ \emph {et~al.}(2012)\citenamefont {Seto},
  \citenamefont {G{\"u}neralp},\ and\ \citenamefont
  {Hutyra}}]{seto_global_2012}%
  \BibitemOpen
  \bibfield  {author} {\bibinfo {author} {\bibfnamefont {K.~C.}\ \bibnamefont
  {Seto}}, \bibinfo {author} {\bibfnamefont {B.}~\bibnamefont {G{\"u}neralp}},
  \ and\ \bibinfo {author} {\bibfnamefont {L.~R.}\ \bibnamefont {Hutyra}},\
  }\href {\doibase 10.1073/pnas.1211658109} {\bibfield  {journal} {\bibinfo
  {journal} {Proceedings of the National Academy of Sciences}\ }\textbf
  {\bibinfo {volume} {109}},\ \bibinfo {pages} {16083} (\bibinfo {year}
  {2012})}\BibitemShut {NoStop}%
\bibitem [{\citenamefont {{United Nations Department of Economic and Social
  Affairs, Population Division}}(2019)}]{unWUP2018}%
  \BibitemOpen
  \bibfield  {author} {\bibinfo {author} {\bibnamefont {{United Nations
  Department of Economic and Social Affairs, Population Division}}},\
  }\href@noop {} {\emph {\bibinfo {title} {World Urbanization Prospects, the
  2018 Revision}}}\ (\bibinfo  {publisher} {{United Nations}},\ \bibinfo
  {address} {{New York}},\ \bibinfo {year} {2019})\BibitemShut {NoStop}%
\bibitem [{\citenamefont {Prein}\ \emph {et~al.}(2017)\citenamefont {Prein},
  \citenamefont {Rasmussen}, \citenamefont {Ikeda}, \citenamefont {Liu},
  \citenamefont {Clark},\ and\ \citenamefont
  {Holland}}]{PrecipitationExtremes}%
  \BibitemOpen
  \bibfield  {author} {\bibinfo {author} {\bibfnamefont {A.~F.}\ \bibnamefont
  {Prein}}, \bibinfo {author} {\bibfnamefont {R.~M.}\ \bibnamefont
  {Rasmussen}}, \bibinfo {author} {\bibfnamefont {K.}~\bibnamefont {Ikeda}},
  \bibinfo {author} {\bibfnamefont {C.}~\bibnamefont {Liu}}, \bibinfo {author}
  {\bibfnamefont {M.~P.}\ \bibnamefont {Clark}}, \ and\ \bibinfo {author}
  {\bibfnamefont {G.~J.}\ \bibnamefont {Holland}},\ }\href {\doibase
  10.1038/nclimate3168} {\bibfield  {journal} {\bibinfo  {journal} {Nature
  Climate Change}\ }\textbf {\bibinfo {volume} {7}},\ \bibinfo {pages} {48}
  (\bibinfo {year} {2017})}\BibitemShut {NoStop}%
\bibitem [{\citenamefont {Huang}\ \emph {et~al.}(2016)\citenamefont {Huang},
  \citenamefont {Yu}, \citenamefont {Guan}, \citenamefont {Wang},\ and\
  \citenamefont {Guo}}]{DrylandExpansion}%
  \BibitemOpen
  \bibfield  {author} {\bibinfo {author} {\bibfnamefont {J.}~\bibnamefont
  {Huang}}, \bibinfo {author} {\bibfnamefont {H.}~\bibnamefont {Yu}}, \bibinfo
  {author} {\bibfnamefont {X.}~\bibnamefont {Guan}}, \bibinfo {author}
  {\bibfnamefont {G.}~\bibnamefont {Wang}}, \ and\ \bibinfo {author}
  {\bibfnamefont {R.}~\bibnamefont {Guo}},\ }\href {\doibase
  10.1038/nclimate2837} {\bibfield  {journal} {\bibinfo  {journal} {Nature
  Climate Change}\ }\textbf {\bibinfo {volume} {6}},\ \bibinfo {pages} {166}
  (\bibinfo {year} {2016})}\BibitemShut {NoStop}%
\bibitem [{\citenamefont {{van de Ven}}\ \emph {et~al.}(2021)\citenamefont
  {{van de Ven}}, \citenamefont {{Capellan-Per{\'e}z}}, \citenamefont {Arto},
  \citenamefont {Cazcarro}, \citenamefont {{de Castro}}, \citenamefont
  {Patel},\ and\ \citenamefont
  {{Gonzalez-Eguino}}}]{van_de_ven_potential_2021}%
  \BibitemOpen
  \bibfield  {author} {\bibinfo {author} {\bibfnamefont {D.-J.}\ \bibnamefont
  {{van de Ven}}}, \bibinfo {author} {\bibfnamefont {I.}~\bibnamefont
  {{Capellan-Per{\'e}z}}}, \bibinfo {author} {\bibfnamefont {I.}~\bibnamefont
  {Arto}}, \bibinfo {author} {\bibfnamefont {I.}~\bibnamefont {Cazcarro}},
  \bibinfo {author} {\bibfnamefont {C.}~\bibnamefont {{de Castro}}}, \bibinfo
  {author} {\bibfnamefont {P.}~\bibnamefont {Patel}}, \ and\ \bibinfo {author}
  {\bibfnamefont {M.}~\bibnamefont {{Gonzalez-Eguino}}},\ }\href {\doibase
  10.1038/s41598-021-82042-5} {\bibfield  {journal} {\bibinfo  {journal} {Sci
  Rep}\ }\textbf {\bibinfo {volume} {11}},\ \bibinfo {pages} {2907} (\bibinfo
  {year} {2021})}\BibitemShut {NoStop}%
\bibitem [{\citenamefont {Bibri}\ \emph {et~al.}(2020)\citenamefont {Bibri},
  \citenamefont {Krogstie},\ and\ \citenamefont
  {K{\"a}rrholm}}]{bibri_compact_2020}%
  \BibitemOpen
  \bibfield  {author} {\bibinfo {author} {\bibfnamefont {S.~E.}\ \bibnamefont
  {Bibri}}, \bibinfo {author} {\bibfnamefont {J.}~\bibnamefont {Krogstie}}, \
  and\ \bibinfo {author} {\bibfnamefont {M.}~\bibnamefont {K{\"a}rrholm}},\
  }\href {\doibase 10.1016/j.dibe.2020.100021} {\bibfield  {journal} {\bibinfo
  {journal} {Developments in the Built Environment}\ }\textbf {\bibinfo
  {volume} {4}},\ \bibinfo {pages} {100021} (\bibinfo {year}
  {2020})}\BibitemShut {NoStop}%
\bibitem [{\citenamefont {Anguelovski}\ \emph {et~al.}(2016)\citenamefont
  {Anguelovski}, \citenamefont {Shi}, \citenamefont {Chu}, \citenamefont
  {Gallagher}, \citenamefont {Goh}, \citenamefont {Lamb}, \citenamefont
  {Reeve},\ and\ \citenamefont {Teicher}}]{anguelovski_equity_2016}%
  \BibitemOpen
  \bibfield  {author} {\bibinfo {author} {\bibfnamefont {I.}~\bibnamefont
  {Anguelovski}}, \bibinfo {author} {\bibfnamefont {L.}~\bibnamefont {Shi}},
  \bibinfo {author} {\bibfnamefont {E.}~\bibnamefont {Chu}}, \bibinfo {author}
  {\bibfnamefont {D.}~\bibnamefont {Gallagher}}, \bibinfo {author}
  {\bibfnamefont {K.}~\bibnamefont {Goh}}, \bibinfo {author} {\bibfnamefont
  {Z.}~\bibnamefont {Lamb}}, \bibinfo {author} {\bibfnamefont {K.}~\bibnamefont
  {Reeve}}, \ and\ \bibinfo {author} {\bibfnamefont {H.}~\bibnamefont
  {Teicher}},\ }\href {\doibase 10.1177/0739456X16645166} {\bibfield  {journal}
  {\bibinfo  {journal} {Journal of Planning Education and Research}\ }\textbf
  {\bibinfo {volume} {36}},\ \bibinfo {pages} {333} (\bibinfo {year}
  {2016})}\BibitemShut {NoStop}%
\bibitem [{\citenamefont {Rahman}\ and\ \citenamefont
  {Szab{\'o}}(2021)}]{MOLA-Review}%
  \BibitemOpen
  \bibfield  {author} {\bibinfo {author} {\bibfnamefont {M.~M.}\ \bibnamefont
  {Rahman}}\ and\ \bibinfo {author} {\bibfnamefont {G.}~\bibnamefont
  {Szab{\'o}}},\ }\href {\doibase 10.1016/j.scs.2021.103214} {\bibfield
  {journal} {\bibinfo  {journal} {Sustainable Cities and Society}\ }\textbf
  {\bibinfo {volume} {74}},\ \bibinfo {pages} {103214} (\bibinfo {year}
  {2021})}\BibitemShut {NoStop}%
\bibitem [{\citenamefont {Liu}\ \emph {et~al.}(2016)\citenamefont {Liu},
  \citenamefont {Peng}, \citenamefont {Jiao},\ and\ \citenamefont
  {Liu}}]{PS-MOLA}%
  \BibitemOpen
  \bibfield  {author} {\bibinfo {author} {\bibfnamefont {Y.}~\bibnamefont
  {Liu}}, \bibinfo {author} {\bibfnamefont {J.}~\bibnamefont {Peng}}, \bibinfo
  {author} {\bibfnamefont {L.}~\bibnamefont {Jiao}}, \ and\ \bibinfo {author}
  {\bibfnamefont {Y.}~\bibnamefont {Liu}},\ }\href {\doibase
  10.1371/journal.pone.0157728} {\bibfield  {journal} {\bibinfo  {journal}
  {PLOS ONE}\ }\textbf {\bibinfo {volume} {11}},\ \bibinfo {pages} {e0157728}
  (\bibinfo {year} {2016})}\BibitemShut {NoStop}%
\bibitem [{\citenamefont {Song}\ and\ \citenamefont
  {Chen}(2018{\natexlab{a}})}]{NGSA-MOLA}%
  \BibitemOpen
  \bibfield  {author} {\bibinfo {author} {\bibfnamefont {M.}~\bibnamefont
  {Song}}\ and\ \bibinfo {author} {\bibfnamefont {D.}~\bibnamefont {Chen}},\
  }\href {\doibase 10.1080/10095020.2018.1489576} {\bibfield  {journal}
  {\bibinfo  {journal} {Geo-spatial Information Science}\ }\textbf {\bibinfo
  {volume} {21}},\ \bibinfo {pages} {273} (\bibinfo {year}
  {2018}{\natexlab{a}})}\BibitemShut {NoStop}%
\bibitem [{\citenamefont {Stewart}\ \emph {et~al.}(2004)\citenamefont
  {Stewart}, \citenamefont {Janssen},\ and\ \citenamefont {{van
  Herwijnen}}}]{GA-MOLA-2004}%
  \BibitemOpen
  \bibfield  {author} {\bibinfo {author} {\bibfnamefont {T.~J.}\ \bibnamefont
  {Stewart}}, \bibinfo {author} {\bibfnamefont {R.}~\bibnamefont {Janssen}}, \
  and\ \bibinfo {author} {\bibfnamefont {M.}~\bibnamefont {{van Herwijnen}}},\
  }\href {\doibase 10.1016/S0305-0548(03)00188-6} {\bibfield  {journal}
  {\bibinfo  {journal} {Computers \& Operations Research}\ }\textbf {\bibinfo
  {volume} {31}},\ \bibinfo {pages} {2293} (\bibinfo {year}
  {2004})}\BibitemShut {NoStop}%
\bibitem [{\citenamefont {Zhang}\ \emph {et~al.}(2016)\citenamefont {Zhang},
  \citenamefont {Zeng}, \citenamefont {Jin}, \citenamefont {Shu}, \citenamefont
  {Zhou},\ and\ \citenamefont {Yang}}]{MA-PS-MOLA}%
  \BibitemOpen
  \bibfield  {author} {\bibinfo {author} {\bibfnamefont {H.}~\bibnamefont
  {Zhang}}, \bibinfo {author} {\bibfnamefont {Y.}~\bibnamefont {Zeng}},
  \bibinfo {author} {\bibfnamefont {X.}~\bibnamefont {Jin}}, \bibinfo {author}
  {\bibfnamefont {B.}~\bibnamefont {Shu}}, \bibinfo {author} {\bibfnamefont
  {Y.}~\bibnamefont {Zhou}}, \ and\ \bibinfo {author} {\bibfnamefont
  {X.}~\bibnamefont {Yang}},\ }\href {\doibase 10.1016/j.ecolmodel.2015.10.017}
  {\bibfield  {journal} {\bibinfo  {journal} {Ecological Modelling}\ }\textbf
  {\bibinfo {volume} {320}},\ \bibinfo {pages} {334} (\bibinfo {year}
  {2016})}\BibitemShut {NoStop}%
\bibitem [{\citenamefont {Mohammadi}\ \emph {et~al.}(2015)\citenamefont
  {Mohammadi}, \citenamefont {Nastaran},\ and\ \citenamefont
  {Sahebgharani}}]{NGSAII-MOLA-2015}%
  \BibitemOpen
  \bibfield  {author} {\bibinfo {author} {\bibfnamefont {M.}~\bibnamefont
  {Mohammadi}}, \bibinfo {author} {\bibfnamefont {M.}~\bibnamefont {Nastaran}},
  \ and\ \bibinfo {author} {\bibfnamefont {A.}~\bibnamefont {Sahebgharani}},\
  }\href {\doibase 10.17485/ijst/2015/v8iS3/60700} {\bibfield  {journal}
  {\bibinfo  {journal} {INDJST}\ }\textbf {\bibinfo {volume} {8}},\ \bibinfo
  {pages} {1} (\bibinfo {year} {2015})}\BibitemShut {NoStop}%
\bibitem [{\citenamefont {Liu}\ \emph {et~al.}(2012)\citenamefont {Liu},
  \citenamefont {Li}, \citenamefont {Shi}, \citenamefont {Huang},\ and\
  \citenamefont {Liu}}]{MACO-MOLA}%
  \BibitemOpen
  \bibfield  {author} {\bibinfo {author} {\bibfnamefont {X.}~\bibnamefont
  {Liu}}, \bibinfo {author} {\bibfnamefont {X.}~\bibnamefont {Li}}, \bibinfo
  {author} {\bibfnamefont {X.}~\bibnamefont {Shi}}, \bibinfo {author}
  {\bibfnamefont {K.}~\bibnamefont {Huang}}, \ and\ \bibinfo {author}
  {\bibfnamefont {Y.}~\bibnamefont {Liu}},\ }\href {\doibase
  10.1080/13658816.2011.635594} {\bibfield  {journal} {\bibinfo  {journal}
  {International Journal of Geographical Information Science}\ }\textbf
  {\bibinfo {volume} {26}},\ \bibinfo {pages} {1325} (\bibinfo {year}
  {2012})}\BibitemShut {NoStop}%
\bibitem [{\citenamefont {Gao}\ \emph {et~al.}(2021)\citenamefont {Gao},
  \citenamefont {Wang}, \citenamefont {Cushman}, \citenamefont {Cheng},
  \citenamefont {Song},\ and\ \citenamefont {Ye}}]{MOLA-SDG}%
  \BibitemOpen
  \bibfield  {author} {\bibinfo {author} {\bibfnamefont {P.}~\bibnamefont
  {Gao}}, \bibinfo {author} {\bibfnamefont {H.}~\bibnamefont {Wang}}, \bibinfo
  {author} {\bibfnamefont {S.~A.}\ \bibnamefont {Cushman}}, \bibinfo {author}
  {\bibfnamefont {C.}~\bibnamefont {Cheng}}, \bibinfo {author} {\bibfnamefont
  {C.}~\bibnamefont {Song}}, \ and\ \bibinfo {author} {\bibfnamefont
  {S.}~\bibnamefont {Ye}},\ }\href {\doibase 10.1007/s10980-020-01051-3}
  {\bibfield  {journal} {\bibinfo  {journal} {Landscape Ecol}\ }\textbf
  {\bibinfo {volume} {36}},\ \bibinfo {pages} {1877} (\bibinfo {year}
  {2021})}\BibitemShut {NoStop}%
\bibitem [{\citenamefont {Song}\ and\ \citenamefont
  {Chen}(2018{\natexlab{b}})}]{SongChenMOLA}%
  \BibitemOpen
  \bibfield  {author} {\bibinfo {author} {\bibfnamefont {M.}~\bibnamefont
  {Song}}\ and\ \bibinfo {author} {\bibfnamefont {D.}~\bibnamefont {Chen}},\
  }\href {\doibase 10.1080/19475683.2018.1424736} {\bibfield  {journal}
  {\bibinfo  {journal} {Annals of GIS}\ }\textbf {\bibinfo {volume} {24}},\
  \bibinfo {pages} {19} (\bibinfo {year} {2018}{\natexlab{b}})}\BibitemShut
  {NoStop}%
\bibitem [{\citenamefont {{van Anders}}\ \emph {et~al.}(2014)\citenamefont
  {{van Anders}}, \citenamefont {Klotsa}, \citenamefont {Ahmed}, \citenamefont
  {Engel},\ and\ \citenamefont {Glotzer}}]{entint}%
  \BibitemOpen
  \bibfield  {author} {\bibinfo {author} {\bibfnamefont {G.}~\bibnamefont {{van
  Anders}}}, \bibinfo {author} {\bibfnamefont {D.}~\bibnamefont {Klotsa}},
  \bibinfo {author} {\bibfnamefont {N.~K.}\ \bibnamefont {Ahmed}}, \bibinfo
  {author} {\bibfnamefont {M.}~\bibnamefont {Engel}}, \ and\ \bibinfo {author}
  {\bibfnamefont {S.~C.}\ \bibnamefont {Glotzer}},\ }\href {\doibase
  10.1073/pnas.1418159111} {\bibfield  {journal} {\bibinfo  {journal} {Proc.
  Natl. Acad. Sci. U.S.A.}\ }\textbf {\bibinfo {volume} {111}},\ \bibinfo
  {pages} {E4812} (\bibinfo {year} {2014})}\BibitemShut {NoStop}%
\bibitem [{\citenamefont {Goldenfeld}(1992)}]{goldenfeld}%
  \BibitemOpen
  \bibfield  {author} {\bibinfo {author} {\bibfnamefont {N.}~\bibnamefont
  {Goldenfeld}},\ }\href@noop {} {\emph {\bibinfo {title} {Lectures on Phase
  Transitions and the Renormalization Group}}}\ (\bibinfo  {publisher}
  {{Addison-Wesley}},\ \bibinfo {address} {{Reading MA}},\ \bibinfo {year}
  {1992})\BibitemShut {NoStop}%
\bibitem [{\citenamefont {Debenedetti}(1996)}]{debenedetti_metastable_1996}%
  \BibitemOpen
  \bibfield  {author} {\bibinfo {author} {\bibfnamefont {P.~G.}\ \bibnamefont
  {Debenedetti}},\ }\href@noop {} {\emph {\bibinfo {title} {Metastable Liquids:
  Concepts and Principles}}},\ Physical Chemistry\ (\bibinfo  {publisher}
  {{Princeton University Press}},\ \bibinfo {address} {{Princeton, N.J}},\
  \bibinfo {year} {1996})\BibitemShut {NoStop}%
\bibitem [{\citenamefont {Chikazumi}(2009)}]{chikazumi_physics_2009}%
  \BibitemOpen
  \bibfield  {author} {\bibinfo {author} {\bibfnamefont {S.}~\bibnamefont
  {Chikazumi}},\ }\href@noop {} {\emph {\bibinfo {title} {Physics of
  {{Ferromagnetism}}}}},\ \bibinfo {edition} {second edition, second edition}\
  ed.,\ International {{Series}} of {{Monographs}} on {{Physics}}\ (\bibinfo
  {publisher} {{Oxford University Press}},\ \bibinfo {address} {{Oxford, New
  York}},\ \bibinfo {year} {2009})\BibitemShut {NoStop}%
\bibitem [{\citenamefont {Sethna}(2021)}]{Sethna2021}%
  \BibitemOpen
  \bibfield  {author} {\bibinfo {author} {\bibfnamefont {J.}~\bibnamefont
  {Sethna}},\ }\href@noop {} {\emph {\bibinfo {title} {Statistical Mechanics:
  Entropy, Order Parameters, and Complexity}}}\ (\bibinfo  {publisher} {{Oxford
  University Press, USA}},\ \bibinfo {year} {2021})\BibitemShut {NoStop}%
\bibitem [{\citenamefont {Berkes}\ \emph {et~al.}(2000)\citenamefont {Berkes},
  \citenamefont {Folke},\ and\ \citenamefont {Colding}}]{berkes_linking_2000}%
  \BibitemOpen
  \bibfield  {author} {\bibinfo {author} {\bibfnamefont {F.}~\bibnamefont
  {Berkes}}, \bibinfo {author} {\bibfnamefont {C.}~\bibnamefont {Folke}}, \
  and\ \bibinfo {author} {\bibfnamefont {J.}~\bibnamefont {Colding}},\
  }\href@noop {} {\emph {\bibinfo {title} {Linking {{Social}} and {{Ecological
  Systems}}: {{Management Practices}} and {{Social Mechanisms}} for {{Building
  Resilience}}}}}\ (\bibinfo  {publisher} {{Cambridge University Press}},\
  \bibinfo {year} {2000})\BibitemShut {NoStop}%
\bibitem [{\citenamefont {DeFries}\ \emph {et~al.}(2004)\citenamefont
  {DeFries}, \citenamefont {Foley},\ and\ \citenamefont
  {Asner}}]{defries_land-use_2004}%
  \BibitemOpen
  \bibfield  {author} {\bibinfo {author} {\bibfnamefont {R.~S.}\ \bibnamefont
  {DeFries}}, \bibinfo {author} {\bibfnamefont {J.~A.}\ \bibnamefont {Foley}},
  \ and\ \bibinfo {author} {\bibfnamefont {G.~P.}\ \bibnamefont {Asner}},\
  }\href {\doibase 10.1890/1540-9295(2004)002[0249:LCBHNA]2.0.CO;2} {\bibfield
  {journal} {\bibinfo  {journal} {Frontiers in Ecology and the Environment}\
  }\textbf {\bibinfo {volume} {2}},\ \bibinfo {pages} {249} (\bibinfo {year}
  {2004})}\BibitemShut {NoStop}%
\bibitem [{\citenamefont {Clark}\ \emph {et~al.}(2016)\citenamefont {Clark},
  \citenamefont {{van Kerkhoff}}, \citenamefont {Lebel},\ and\ \citenamefont
  {Gallopin}}]{clark_crafting_2016}%
  \BibitemOpen
  \bibfield  {author} {\bibinfo {author} {\bibfnamefont {W.~C.}\ \bibnamefont
  {Clark}}, \bibinfo {author} {\bibfnamefont {L.}~\bibnamefont {{van
  Kerkhoff}}}, \bibinfo {author} {\bibfnamefont {L.}~\bibnamefont {Lebel}}, \
  and\ \bibinfo {author} {\bibfnamefont {G.~C.}\ \bibnamefont {Gallopin}},\
  }\href {\doibase 10.1073/pnas.1601266113} {\bibfield  {journal} {\bibinfo
  {journal} {Proceedings of the National Academy of Sciences}\ }\textbf
  {\bibinfo {volume} {113}},\ \bibinfo {pages} {4570} (\bibinfo {year}
  {2016})}\BibitemShut {NoStop}%
\bibitem [{\citenamefont {Klishin}\ \emph {et~al.}(2018)\citenamefont
  {Klishin}, \citenamefont {Shields}, \citenamefont {Singer},\ and\
  \citenamefont {{van Anders}}}]{systemphys}%
  \BibitemOpen
  \bibfield  {author} {\bibinfo {author} {\bibfnamefont {A.~A.}\ \bibnamefont
  {Klishin}}, \bibinfo {author} {\bibfnamefont {C.~P.}\ \bibnamefont
  {Shields}}, \bibinfo {author} {\bibfnamefont {D.~J.}\ \bibnamefont {Singer}},
  \ and\ \bibinfo {author} {\bibfnamefont {G.}~\bibnamefont {{van Anders}}},\
  }\href {\doibase 10.1088/1367-2630/aae72a} {\bibfield  {journal} {\bibinfo
  {journal} {New J. Phys.}\ }\textbf {\bibinfo {volume} {20}},\ \bibinfo
  {pages} {103038} (\bibinfo {year} {2018})},\ \Eprint
  {http://arxiv.org/abs/1709.03388} {arxiv:1709.03388 [physics.soc-ph]}
  \BibitemShut {NoStop}%
\bibitem [{\citenamefont {Kirkpatrick}\ \emph {et~al.}(1983)\citenamefont
  {Kirkpatrick}, \citenamefont {Gelatt},\ and\ \citenamefont
  {Vecchi}}]{kirkpatrick1983}%
  \BibitemOpen
  \bibfield  {author} {\bibinfo {author} {\bibfnamefont {S.}~\bibnamefont
  {Kirkpatrick}}, \bibinfo {author} {\bibfnamefont {C.~D.}\ \bibnamefont
  {Gelatt}}, \ and\ \bibinfo {author} {\bibfnamefont {M.~P.}\ \bibnamefont
  {Vecchi}},\ }\href {\doibase 10.1126/science.220.4598.671} {\bibfield
  {journal} {\bibinfo  {journal} {Science}\ }\textbf {\bibinfo {volume}
  {220}},\ \bibinfo {pages} {671} (\bibinfo {year} {1983})}\BibitemShut
  {NoStop}%
\bibitem [{\citenamefont {Wolff}(1989)}]{wolff}%
  \BibitemOpen
  \bibfield  {author} {\bibinfo {author} {\bibfnamefont {U.}~\bibnamefont
  {Wolff}},\ }\href {\doibase 10.1103/PhysRevLett.62.361} {\bibfield  {journal}
  {\bibinfo  {journal} {Phys. Rev. Lett.}\ }\textbf {\bibinfo {volume} {62}},\
  \bibinfo {pages} {361} (\bibinfo {year} {1989})}\BibitemShut {NoStop}%
\bibitem [{\citenamefont {{Kent-Dobias}}\ and\ \citenamefont
  {Sethna}(2018)}]{GhostWolff}%
  \BibitemOpen
  \bibfield  {author} {\bibinfo {author} {\bibfnamefont {J.}~\bibnamefont
  {{Kent-Dobias}}}\ and\ \bibinfo {author} {\bibfnamefont {J.~P.}\ \bibnamefont
  {Sethna}},\ }\href {\doibase 10.1103/PhysRevE.98.063306} {\bibfield
  {journal} {\bibinfo  {journal} {Phys. Rev. E}\ }\textbf {\bibinfo {volume}
  {98}},\ \bibinfo {pages} {063306} (\bibinfo {year} {2018})}\BibitemShut
  {NoStop}%
\bibitem [{\citenamefont {Connolly}\ \emph {et~al.}(2023)\citenamefont
  {Connolly}, \citenamefont {Beckett}, \citenamefont {Aliahmadi},\ and\
  \citenamefont {{van Anders}}}]{connollyFlashWolff2023}%
  \BibitemOpen
  \bibfield  {author} {\bibinfo {author} {\bibfnamefont {S.}~\bibnamefont
  {Connolly}}, \bibinfo {author} {\bibfnamefont {M.}~\bibnamefont {Beckett}},
  \bibinfo {author} {\bibfnamefont {H.}~\bibnamefont {Aliahmadi}}, \ and\
  \bibinfo {author} {\bibfnamefont {G.}~\bibnamefont {{van Anders}}},\ }\href
  {\doibase 10.5281/ZENODO.7846107} {\  (\bibinfo {year} {2023}),\
  10.5281/ZENODO.7846107}\BibitemShut {NoStop}%
\bibitem [{\citenamefont {Aliahmadi}\ \emph {et~al.}(2023)\citenamefont
  {Aliahmadi}, \citenamefont {Beckett}, \citenamefont {Connolly}, \citenamefont
  {Chen},\ and\ \citenamefont {{van Anders}}}]{aliahmadi_land_2023}%
  \BibitemOpen
  \bibfield  {author} {\bibinfo {author} {\bibfnamefont {H.}~\bibnamefont
  {Aliahmadi}}, \bibinfo {author} {\bibfnamefont {M.}~\bibnamefont {Beckett}},
  \bibinfo {author} {\bibfnamefont {S.}~\bibnamefont {Connolly}}, \bibinfo
  {author} {\bibfnamefont {D.}~\bibnamefont {Chen}}, \ and\ \bibinfo {author}
  {\bibfnamefont {G.}~\bibnamefont {{van Anders}}},\ }\href {\doibase
  10.5683/SP3/KVXIMS} {\  (\bibinfo {year} {2023}),\
  10.5683/SP3/KVXIMS}\BibitemShut {NoStop}%
\end{thebibliography}
\end{document}